\documentclass[compactaffiliation]{Interspeech}

\interspeechcameraready

\title{Towards Inclusive ASR: Investigating Voice Conversion \\ for Dysarthric Speech Recognition in Low-Resource Languages}

\author[affiliation={1},equalcontribution]{Chin-Jou}{Li}
\author[affiliation={1},equalcontribution]{Eunjung}{Yeo}
\author[affiliation={1}]{Kwanghee}{Choi}
\author[affiliation={2,3}]{Paula Andrea}{Pérez-Toro}
\author[affiliation={1}]{Masao}{Someki}
\author[affiliation={5}]{Rohan Kumar}{Das}
\author[affiliation={4}]{Zhengjun}{Yue}
\author[affiliation={2,3}]{Juan Rafael}{Orozco-Arroyave}
\author[affiliation={2}]{Elmar}{Nöth}
\author[affiliation={1}]{David R.}{Mortensen}

\affiliation{}{Carnegie Mellon University}{USA}
\affiliation{}{FAU Erlangen-Nürnberg}{Germany}
\affiliation{}{Universidad de Antioquia}{Colombia}
\affiliation{}{TU Delft}{the Netherlands}
\affiliation{}{Fortemedia}{Singapore}
\email{\{chinjoul,eyeo2,dmortens\}@andrew.cmu.edu}

\usepackage{comment}
\usepackage{hyperref}
\usepackage{cleveref}
\usepackage{multirow}
\usepackage{cite}
\usepackage{subfigure}
\usepackage{tabularray}
\usepackage{booktabs}
\usepackage{pifont}% http://ctan.org/pkg/pifont
\newcommand{\xmark}{\ding{55}}%

\begin{document}
\maketitle

% the abstract here must exactly match the abstract entered into the paper submission system

\begin{abstract} % 963/1000 characters
Automatic speech recognition (ASR) for dysarthric speech remains challenging due to data scarcity, particularly in non-English languages. To address this, we fine-tune a voice conversion model on English dysarthric speech (UASpeech) to encode both speaker characteristics and prosodic distortions, then apply it to convert healthy non-English speech (FLEURS) into non-English dysarthric-like speech. The generated data is then used to fine-tune a multilingual ASR model, Massively Multilingual Speech (MMS), for improved dysarthric speech recognition. Evaluation on PC-GITA (Spanish), EasyCall (Italian), and SSNCE (Tamil) demonstrates that VC with both speaker and prosody conversion significantly outperforms the off-the-shelf MMS performance and conventional augmentation techniques such as speed and tempo perturbation. Objective and subjective analyses of the generated data further confirm that the generated speech simulates dysarthric characteristics.
\end{abstract}

\section{Introduction}

Dysarthria is a motor speech disorder caused by neurological conditions such as Cerebral Palsy (CP), Parkinson's disease (PD), and Amyotrophic Lateral Sclerosis (ALS) \cite{simmons1997use}.
These neurological impairments affect the coordination and strength of the muscles involved in speech production, resulting in reduced speech intelligibility \cite{enderby1980frenchay}.
Consequently, despite notable progress, automatic speech recognition (ASR) systems still struggle to process dysarthric speech \cite{hasegawa2024community}.

A major challenge in improving ASR for dysarthric speech is the scarcity of annotated dysarthric speech datasets \cite{zheng2023improving, yeo2025potential}. 
Collecting such data is inherently difficult, as recording sessions can be physically demanding for individuals with dysarthria, resulting in limited availability of large, high-quality corpora. Currently, there are only around ten publicly available dysarthric speech datasets \cite{bhat2025speech}, with four of them covering English \cite{menendez1996nemours, kim2008dysarthric, rudzicz2012torgo, hasegawa2024community}. The situation is even more critical for non-English languages, where dysarthric speech resources are largely nonexistent. This severe data imbalance restricts the development of robust ASR models for non-English dysarthric speech, exacerbating accessibility challenges for individuals with dysarthria across diverse linguistic communities.

To address data scarcity, researchers have explored data augmentation techniques such as vocal tract length perturbation (VTLP) \cite{geng20_interspeech, karumuru2024domain}, pitch and tempo modification \cite{vachhani2018data, geng20_interspeech, jiao2018simulating}, speech rate adjustments \cite{wang2024enhancing, yue2022raw}, and formant transformations \cite{karumuru2024domain}. 
More recently, text-to-speech (TTS) \cite{soleymanpour2024accurate, hermann2023few} and voice conversion (VC) \cite{jiao2018simulating, vachhani2018data, Rohan2021significance, zheng2023improving} has emerged as a promising strategy for synthesizing dysarthric speech data. 
Unlike traditional perturbation methods that manipulate isolated acoustic features, VC offers a more comprehensive style transfer by transforming healthy speech to exhibit the acoustic characteristics of dysarthric speech \cite{yang2021cross}.

Although VC-based methods for generating dysarthric-like speech have shown promise, they have primarily been applied in scenarios where data in the target language are available \cite{jiao2018simulating, zheng2023improving}. However, this assumption does not hold for most languages, where dysarthric speech corpora are absent. Addressing this gap is crucial for developing {\it inclusive ASR} systems that support diverse linguistic populations. The application of learned transformations from one language to another as a data augmentation strategy has been explored in other speech processing tasks, such as child speech ASR \cite{zhang2024improving} and non-native speech ASR \cite{zhang2023exploring}. These studies have demonstrated that VC-based approaches outperform traditional augmentation techniques, substantially improving ASR performance.

% This suggests that considering VC technologies could be a promising direction for improving dysarthric speech ASR in low-resource languages.

Building upon these insights, this study investigates a VC-based style transfer approach for generating dysarthric-like speech for languages lacking dysarthric data. Specifically, we leverage English dysarthric speech to capture the acoustic and prosodic markers of dysarthria, applying these learned transformations to healthy speech in other languages. We evaluate our method for dysarthric ASR in Spanish, Italian, and Tamil, comparing its performance against an off-the-shelf multilingual ASR model and fine-tuned models incorporating conventional augmentations. In addition, we perform objective and subjective evaluations of the generated speech to assess whether the generated data reflects the characteristics of dysarthric speech.

To the best of our knowledge, this is the first attempt to generate dysarthric speech datasets in a setting where no such data is available for the target language. 
By bridging the data gap between high- and low-resource languages in terms of dysarthria, this approach contributes to developing more inclusive ASR technologies, ultimately improving accessibility for individuals with dysarthria across underrepresented linguistic contexts.

\section{Method}\label{sec:datasets}
\begin{figure}[t]
    \centering
    \includegraphics[width=0.85\columnwidth]{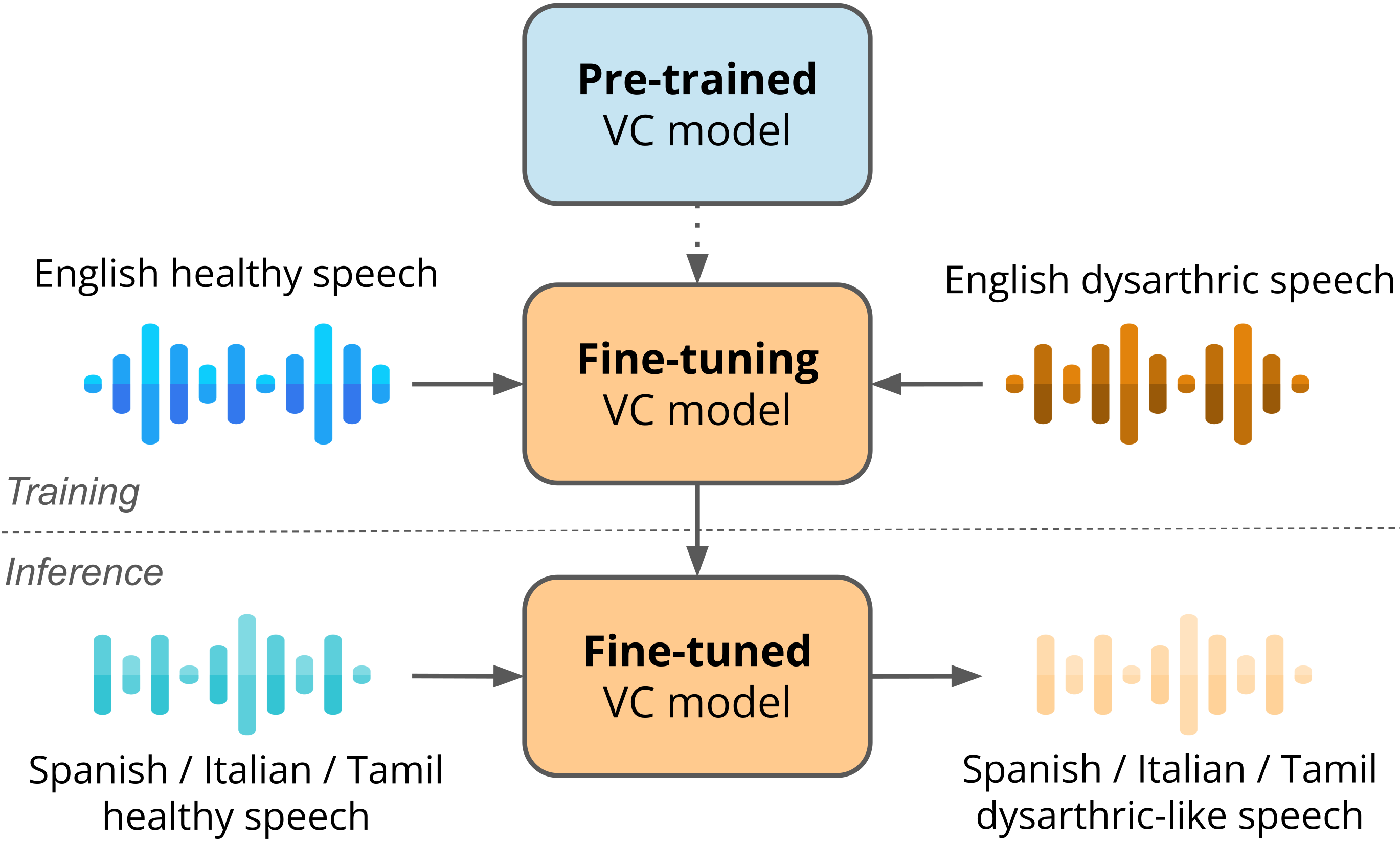}
    \vspace{-2mm}
    \caption{
    Overview of the proposed VC-based framework. The pre-trained VC model is fine-tuned on English healthy and dysarthric speech (UASpeech). During inference, the fine-tuned VC model converts non-English healthy speech (FLEURS) into dysarthric-like speech in the same non-English language.}    
    \vspace{-6mm}
    % English dysarthric speech as a cross-lingual reference.}
    \label{fig:summary}
\end{figure}

{\subsection{Non-English Dysarthric Speech Generation}}
To generate dysarthric-like speech in non-English languages, we first fine-tuned an existing VC model to capture the distinctive acoustic and prosodic characteristics of dysarthric speech using English dysarthric data . 
We then applied the VC model to transform non-English healthy speech into dysarthric-like speech in the same language (\Cref{fig:summary}). 
% An overview of our proposed framework is illustrated in .

In this study, we considered Unified Unsupervised Voice Conversion (UUVC) \cite{chen2023unified} for two primary reasons: (1) its speaker conversion performance is comparable to state-of-the-art VC methods, and (2) it explicitly encodes pitch-energy and rhythm (duration) attributes alongside speaker characteristics. Since irregularities in pitch, energy, and rhythm are defining features of dysarthric speech \cite{simmons1997use, enderby1980frenchay}, transferring these attributes may enhance the realism of the synthesized dysarthric speech.

\subsection{Automatic Speech Recognition}
For the ASR model, we employed Massively Multilingual Speech (MMS) \cite{pratap2023mms}, a self-supervised multilingual model which supports 1162 languages.
We chose a self-supervised model for its strong ability to model phonetic variation even with small data, which is crucial in handling atypical pronunciations \cite{yeo23_interspeech, choi2025leveraging}.
Multilinguality is essential for this study as it allows us to fairly evaluate the off-the-shelf performance of dysarthric speech ASR in different languages and the effectiveness of data augmentation. MMS was chosen for its Connectionist Temporal Classification (CTC) framework, which is more resistant to hallucinations than autoregressive sequence-to-sequence models \cite{peng-etal-2024-owsm}.
As dysarthric speech often deviates from typical speech patterns, the monotonic alignment constraints of CTC help reduce spurious insertions and improve transcription stability.

\section{Experiments}\label{sec:experiments}
\subsection{Datasets}
This study utilizes five datasets: one multilingual speech dataset collected from the general population, and four monolingual dysarthric speech datasets in English, Spanish, Italian, and Tamil. The multilingual dataset serves as healthy speech to be converted, while the English dysarthric dataset—being the most accessible—is used for VC fine-tuning and inference to generate dysarthric-like speech. 
The remaining datasets are for ASR experiments. 
\Cref{tab:dataset} summarizes the datasets used in this study.

\noindent\textbf{FLEURS} \cite{conneau2023fleurs} is a n-way parallel multilingual speech corpus containing read speech from Wikipedia in 102 languages, including around 13 hours of Spanish, 12 hours of Italian, and 14 hours of Tamil speech.
% en: 37120 sec; es: 47697.6 sec; it: 50637.7 sec; ta: 43431.8 sec

\noindent\textbf{UASpeech} \cite{kim2008dysarthric} is an English dysarthric speech dataset that includes 15 individuals with Cerebral Palsy and 13 age-matched healthy controls. The dataset is organized into three blocks, each containing an identical set of 255 words and 100 uncommon words unique to itself. 
%, supplemented by 300 uncommon words evenly distributed across the blocks, each featuring a distinct subset of these uncommon words. 
% The corpus was developed to support research in ASR systems for individuals with speech impairments.

\noindent\textbf{PC-GITA} \cite{orozco2014new} includes 50 patients with Parkinson's Disease (PD) and 50 healthy subjects evenly matched in age and gender, all native Spanish speakers from Colombia. Each participant read 10 phonetically balanced sentences and 25 isolated words representing the phonetic inventory of Colombian Spanish. %Expert neurologists and phoniatricians performed assessments in all PD patients. 

\noindent\textbf{EasyCall} \cite{turrisi2021easycall} is an Italian dysarthric speech dataset comprising 21 healthy speakers and 26 dysarthric speakers.
The dysarthric group includes individuals with conditions such as PD, Huntington's Disease, ALS, peripheral neuropathy, and myopathic or myasthenic lesions. 
Each speaker recorded 66 to 69 mobile commands, including isolated words and short phrases. 
% The corpus was specifically designed to enhance smartphone ASR systems and contains simple mobile command phrases. 
% The individual degree of speech impairment was assessed by neurologists through the Therapy Outcome Measure.

\noindent\textbf{SSNCE} \cite{ta2016dysarthric} consists of Tamil speech recordings from 20 speakers with dysarthria and 10 healthy control subjects. All dysarthric speakers were diagnosed with CP. Each speaker recorded 103 unique words and 262 unique sentences containing 2 to 6 words.

\begin{table}[t!]
\caption{Summary of datasets. EN, ES, IT, and TA refer to English, Spanish, Italian, and Tamil, respectively. Cat. refers to Category, where D and H refer to dysarthria and healthy, respectively.}
\vspace{-2mm}
\label{tab:dataset}
\centering
\resizebox{0.97\linewidth}{!}{%
\begin{tabular}{l l r r r r r}
\toprule
\multirow{2}{*}{\bf Dataset} & \multirow{2}{*}{\bf{Cat.}} & \multirow{2}{*}{\bf \# Spk.} & \multicolumn{2}{c}{\bf \# \bf Utterances} & \multicolumn{2}{c}{\bf Duration (hrs.)} \\
\cmidrule(lr){4-5} \cmidrule(lr){6-7}
& & & Words &  Sent. & Words &  Sent. \\
\midrule
% FLEURS       & \textcolor{blue}{-} & - & 15376 & - & 49.68 \\
% FLEURS (EN)  & \textcolor{blue}{-} & - & 3643 & - & 10.31 \\
FLEURS (ES) & H & ---  & ---   & 4112 & ---    & 13.25 \\
FLEURS (IT) & H & ---  & ---   & 3335 & ---   & 12.06 \\
FLEURS (TA) & H &---  & ---   & 4285 &---    & 14.07 \\
\midrule
\multirow{2}{*}{UASpeech (EN)} & D &15  & 39150 & ---   & 39.59 & --- \\
 &  H & 13 & 104415 &---   & 60.63 & --- \\
\midrule
\multirow{2}{*}{PC-GITA (ES)} & D & 50  & 1249  & 555 & 0.20  & 0.56 \\
  & H & 50  &  1251 &  542  &  0.21 & 0.56 \\
\multirow{2}{*}{EasyCall (IT)} & D  & 26  & 9179 & 2130 & 6.28 & 2.22\\
   & H & 21  & 8214 & 1863 & 3.09 & 0.90 \\
\multirow{2}{*}{SSNCE (TA)}  & D & 20  & 2060  & 5240 & 0.68 & 4.77 \\
   & H & 10 & 182 &  3468  & 0.18 & 2.75 \\
\bottomrule
\end{tabular}
}
\vspace{-5mm}
\end{table}

\begin{table*}[t]
\caption{
ASR performance with fine-tuning data size (Hours) from FLEURS and its augmentation. None refers to fine-tuning on FLEURS without any augmentation method. Word- and sentence-level evaluations use CER (\%) for each language. 
% Lower CER indicates better performance.
%bf results in each setting are \bf{bolded}. \bf{Underline} indicates MAPSSWE significance ($p < 0.05$) against all others but another VC method and \bf{blue} if significant overall.
}
\vspace{-2mm}
\label{tab:main}
\centering
\resizebox{0.95\textwidth}{!}{
\begin{tabular}{lclr rr rr rr rr rr}
\toprule
\multirow{2}{*}{\bf Language} & \multirow{2}{*}{\bf Fine-tune} & \multirow{2}{*}{\bf Augmentation} & \multirow{2}{*}{\bf Hours} & \multicolumn{2}{c}{\bf All} & \multicolumn{2}{c}{\bf Healthy} & \multicolumn{2}{c}{\bf Mild} & \multicolumn{2}{c}{\bf Moderate} & \multicolumn{2}{c}{\bf Severe} \\
\cmidrule(lr){5-6} \cmidrule(lr){7-8} \cmidrule(lr){9-10} \cmidrule(lr){11-12} \cmidrule(lr){13-14}
& & & & Word & Sent. & Word & Sent. & Word & Sent. & Word & Sent. & Word & Sent. \\
\midrule
\multirow{6}{*}{Spanish} 
& \xmark & N/A & --- & 39.9 & 47.9 & 30.5 & 43.7 & 47.1 & 52.6 & 48.5 & 50.4 & 57.5 &  57.7\\
& \checkmark & None & 10.2 & 43.0 & 48.6 & 32.4 & 43.4 & 48.6 & 51.4 & 53.0 & 50.4 & 65.5 & 69.1 \\
& \checkmark & Speed & 20.6 & 44.7 & 47.6 & 35.1 & 43.2 & 50.7 & 50.9 & 53.4 & 49.9 & 64.7 & 60.6 \\
& \checkmark & Tempo  & 20.6 & 51.7 & 48.2 & 40.8 & 43.2 & 54.7 & 50.9 & 61.0 & 51.0 & 81.6 & 64.8 \\
\cmidrule{3-14}
& \checkmark & Speaker-only VC & 20.3 & 34.9 & 47.7 & 25.8 & 43.5 & 40.4 & 52.8 & 41.9 & 51.0 & 57.9 & \bf{54.3} \\
& \checkmark & Speaker-prosody VC  & 20.4 & \bf{27.0} & \bf{46.8} & \bf{18.5} & \bf{42.8} & \bf{30.5} & \bf{49.9} & \bf{34.6} & \bf{50.0} & \bf{47.6} & 55.7 \\

\midrule
\multirow{6}{*}{Italian} 
& \xmark & N/A & --- & 106.6 & 63.1 & 41.3 & 25.0 & 197.4 & 100.2 & 204.2 & 132.3 & 123.0 & 80.1 \\
& \checkmark & None & 10.5 & 60.4 & 42.0 & 30.2 & 18.4 & 94.7 & 62.8 & 99.3 & 75.2 & 100.5 & 79.6 \\
& \checkmark & Speed & 21.4 & 76.8 & 48.6 & 61.6 & 31.8 & 88.5 & 57.9 & 101.4 & 75.9 & 97.5 & 77.5 \\
& \checkmark & Tempo  & 21.4 & 81.8 & 49.1 & 65.7 & 31.8 & 99.8 & 62.1 & 105.9 & 74.7 & 96.4 & 78.3 \\
\cmidrule{3-14}
& \checkmark & Speaker-only VC & 21.1 & 52.3 & 36.2 & \bf{25.8} & \bf{16.6} & 75.1 & 46.1 & 90.2 & 68.6 & 94.6 & 70.2 \\
& \checkmark & Speaker-prosody VC & 22.4 & \bf{48.6} & \bf{34.5} & 26.6 & 16.7 & \bf{68.0} & \bf{42.0} & \bf{77.6} & \bf{64.0} & \bf{88.1} & \bf{69.0} \\

\midrule
\multirow{6}{*}{Tamil}
& \xmark & N/A & --- & 60.1 & 42.1 & 16.0 & 7.8 & 56.4 & 27.7 & 94.4 & 70.0 & 96.5 & 89.8 \\
& \checkmark & None & 9.9 & 56.2 & 42.8 & 14.6 & 8.5 & 47.9 & 30.0 & 89.8 & 70.3 & 97.1 & 88.8 \\
& \checkmark & Speed & 20.1 & 48.1 & 48.3 & 9.7  & 8.8 & 43.0 & 32.5 & 75.3 & 81.4 & 91.9 & 98.9 \\
& \checkmark & Tempo  & 20.1 & 62.6 & 45.8 & 19.9 & 8.6 & 59.2 & 31.4 & 95.2 & 75.5 & 99.4 & 97.5 \\
\cmidrule{3-14}
& \checkmark & Speaker-only VC & 19.9 & 38.2 & 34.7 & \bf{6.4} & \bf{7.5} & 29.7 & \bf{24.9} & 58.6 & 51.4 & 89.1 & 86.0 \\
& \checkmark & Speaker-prosody VC & 20.6 & \bf{34.0} & \bf{34.1} & 6.6 & 7.9 & \bf{23.9} & 25.0 & \bf{51.1} & \bf{50.2} & \bf{85.0} & \bf{82.9} \\
\bottomrule
\end{tabular}
}
\vspace{-4mm}
\end{table*}

\subsection{Voice Conversion System} \label{ssec:cl-vc}
Our proposed VC-based framework for non-English dysarthric speech generation is illustrated in \Cref{fig:summary}.
%As illustrated in \Cref{fig:summary}, our VC method consists of three steps: pretraining, fine-tuning, and inference.
We used the UUVC checkpoint \textit{pre-trained} on LibriTTS \cite{zen19_interspeech}, VCTK \cite{veaux2013voice}, and LJSpeech \cite{ito2017lj}, as provided in the official repository \cite{chen2023unified}. We then \textit{fine-tuned} the model on the entire UASpeech dataset, incorporating both healthy and dysarthric speech, for 10,000 steps. This is to enable the model to learn the acoustic characteristics of dysarthria. The inclusion of healthy speech was intended to stabilize the training process, as the high variability in dysarthric speech alone could lead to unstable model convergence. During \textit{inference}, utterances from the train and validation set of FLEURS served as source utterances for each language. Target utterances were chosen from dysarthric speakers in UASpeech. To address gender imbalance (11 males, 4 females), we upsampled the female speakers twice, creating 19 target utterances. Each source utterance was randomly paired with a target utterance. Since UUVC modifies speech characteristics independently, we first converted speaker identity, then applied prosody modifications. This process resulted in two types of voice-converted speech: speaker-only VC and speaker-prosody VC.

% We generated two types of voice-converted speech: speaker-only VC and speaker-prosody VC. Since UUVC modifies speech characteristics independently, we first converted the speaker identity and then applied prosody modification.

%, resulting in two distinct versions. Speaker-only VC alters only the speaker identity, while speaker-prosody VC incorporates both speaker and prosodic transformations.

% \input{figures/overview}

\subsection{Conventional Data Augmentations}
The dysarthric-like speech generation by our proposed VC framework (\Cref{fig:summary}) is used as a data augmentation technique for fine-tuning ASR models in this study. This evaluation will determine whether the proposed framework can effectively enhance ASR performance for dysarthric speech, contributing to the development of a more inclusive ASR system. Additionally, we plan to compare our approach with conventional data augmentation methods to assess its effectiveness in improving dysarthric speech recognition. 

Following \cite{vachhani2018data, geng20_interspeech, Rohan2021significance}, we tested two conventional data augmentation methods for dysarthric speech: speed perturbation and tempo perturbation. These techniques have been widely used to simulate variations in speech disorders and improve ASR robustness. Each method generated ten variations, covering a spectrum from mild to severe dysarthria. To ensure diversity, each utterance was randomly augmented using one of these variations. Both methods were implemented using SoX \cite{barras2012sox}.

\noindent\textbf{Speed perturbation} is a widely used data augmentation technique for dysarthric and pathological speech \cite{vachhani2018data, geng20_interspeech, yue2022raw, wang2024enhancing}. In this method, both pitch and tempo were modified by resampling the input with a speed ratio $R_s$, defined as the ratio of the new speed to the original speed. In our setup, $\{R_s\}$ varied from 0.75 to 1.25 in steps of 0.05, excluding 1.

\noindent\textbf{Tempo perturbation} is another popular augmentation technique for dysarthric speech \cite{Rohan2021significance, vachhani2018data, geng20_interspeech, jiao2018simulating} that modifies speed without altering pitch using the Waveform Similarity based Overlap-Add (WSOLA) algorithm \cite{verhelst1993overlap}. We set the tempo resampling ratios $\{R_t\}$ same as $\{R_s\}$, ranging from 0.75 to 1.25.

\subsection{Automatic Speech Recognition System}
\label{ssec:asr}
We considered MMS-1B-ALL model \cite{pratap2023mms} for all ASR experiments. The model's adapter was fine-tuned for each language on FLEURS, including augmented versions when available, for five epochs with a 0.0005 learning rate. We evaluated ASR performance across four fine-tuning setups: (1) off-the-shelf without fine-tuning, (2) fine-tuning with healthy speech from FLEURS, (3) fine-tuning with healthy speech and conventional augmentations (speed or tempo perturbation), and (4) fine-tuning with our proposed VC approach (speaker-only VC or speaker-prosody VC). ASR performance was assessed on dysarthric speech datasets described in \Cref{sec:datasets}, using character error rate (CER) as an evalaution metric, with separate evaluations for word-level and sentence-level utterances. See our codebase\footnote{\url{https://github.com/chinjouli/dysaug-vc}} for more details.

\section{Results for ASR Studies}\label{sec:results}

% \subsection{Comparison Across Data Augmentation Methods}\label{ssec:result}

\Cref{tab:main} presents ASR performance across Spanish, Italian, and Tamil. We consider the pre-trained ASR model and then fine-tune it with the FLEURS dataset using various data augmentation methods. Within each language, CER increases with dysarthria severity, consistent with previous findings that ASR performance deteriorates as dysarthria severity increases \cite{tu2016relationship, jaddoh2023interaction, hasegawa2024community}. 
Among the data augmentation techniques tested, the speaker-prosody VC approach generally yielded the lowest CER across all languages and dysarthria severity levels, demonstrating its effectiveness in improving ASR performance for dysarthric speech. It is noted that fine-tuning with VC consistently outperforms conventional augmentation methods in all conditions. In addition, speaker-only VC yields results comparable to speaker-prosody VC in most cases, suggesting that speaker identity transformation alone reflects a critical attribute of dysarthric speech in our proposed framework of style transfer from healthy to dysarthric-like speech.
Nevertheless, incorporating prosodic aspects further enhances performance, particularly for dysarthric speech, underscoring the importance of integrating prosodic characteristics when generating dysarthric-like speech.
At the sentence level, ASR demonstrates smaller improvements compared to the word level. 
We attribute this to the use of UASpeech for encoding dysarthric speech. Since it consists solely of isolated words, the model may have limited exposure to sentence-level dysarthric patterns, restrcting its ability to generalize to longer utterances. 
% Further investigation into the underlying causes of these limited enhancements is necessary.

% \subsection{Comparison Across Languages}\label{ssec:analysis-pairs}
% % \Cref{tab:relative}
% Given structural differences between languages, certain language pairs may exhibit greater transferability in VC. To assess this impact, we compared the relative improvement in ASR performance between the off-the-shelf model and the best-performing VC experiments by language. For word-level ASR, Italian demonstrated the highest improvement (54.4\%), followed by Tamil (43.4\%) and Spanish (32.3\%). A similar trend was observed in sentence-level ASR, where Italian again showed the greatest improvement (45.3\%), followed by Tamil (19.0\%), while Spanish exhibited minimal gains (0.02\%). These variations suggest that certain languages may facilitate more effective dysarthric speech transformation from English and subsequent ASR adaptation. 
% Further investigation is necessary to understand the underlying causes of these differences.

%the word-sentence error trend of es/ta are different - bc Tamil words are longer; another reason might be SSNCE contains unique words and sentences, while in PC-GITA, it's repeating.

% \vspace{-2mm}
\section{Analyses of Generated Dysarthric Data}\label{sec:discussion}
ASR performance improvement in \Cref{sec:results} can attributed to various reasons beyond augmented data being similar to dysarthric speech. To verify whether the generated speech exhibits dysarthric characteristics, we conduct both objective and subjective analyses on the generated data.
Unlike in our ASR studies, where the generated data was used for fine-tuning, this section evaluates it as the test set. We pose the following question to both the model and human evaluators: ``Does the generated speech sound like dysarthria?''

\subsection{Objective Evaluation: Classification}
We train an XGBoost classifier \cite{chen2016xgboost} for each language using prosodic features extracted with DisVoice \cite{vasquez2018towards} to distinguish between dysarthric and healthy speech. The model is trained on dysarthria datasets using an 8:2 group-stratified train-test split. We optimize the classifier through grid search and select the best-performing model.
The selected classifier is then used to evaluate whether the generated speech is perceived as healthy or dysarthric\textsuperscript{1}.
The F1-scores on the test set are as follows: 69.98\% (Spanish-All), 82.16\% (Italian-All), 93.99\% (Tamil-All).

\Cref{tab:ratio} presents the proportion of audio samples classified as dysarthric relative to the total number of samples, evaluated under different augmentation techniques. Results on FLEURS with no augmentation (``None'') serve as the baseline, since our generated data originates from FLEURS. Speed and tempo perturbations had minimal impact, suggesting these methods do not effectively simulate dysarthria, aligning with our ASR results. In contrast, VC-based augmentations led to a substantial increase in the proportion of samples classified as dysarthric, confirming that speaker and prosodic modifications enhance dysarthric-like characteristics. Speaker-prosody VC generally outperformed speaker-only VC, reinforcing the role of prosodic transfer in dysarthria simulation. Notably, Italian samples, including FLEURS, were frequently classified as dysarthric, suggesting potential noise in the FLEURS Italian dataset, warranting further investigation.

% \begin{table}[t]
% \caption{Relative improvement in generated data classified as dysarthria, compared to FLEURS data (\%). Word and Sent. in column indicate authentic dysarthric speech used for classifier training.}
% \label{tab:ratio}
% \centering
% % \resizebox{0.87\linewidth}{!}{%
% \begin{tabular}{clrr}
% \toprule
% Lang. & Aug. Method & Word & Sent. \\
% \midrule
% \multirow{4}{*}{Spanish} % f1: 65.00; 69.09
% % & None       & 13.04 & 50.51 \\
% & Speed       & 20.08 & 4.70 \\
% & Tempo      & 7.67 & 0.43 \\
% & Speaker-only VC  & 44.48  & \textbf{58.12}\\
% & Speaker-prosody VC & \textbf{80.53} & 57.33\\
% \midrule
% \multirow{4}{*}{Italian} % f1: 80.77; 89.61
% % & None       & 76.13 & 89.27\\
% & Speed       & -0.55 & -0.69\\
% & Tempo      & 0.98 & -0.53\\
% & Speaker-only VC  & 2.61 & \textbf{5.31}\\
% & Speaker-prosody VC & \textbf{12.17} & 1.14\\
% \midrule
% \multirow{4}{*}{Tamil} % f1: 100; 96.15
% % & None       & 8.04  & 62.25\\
% & Speed       & -4.19 & 2.10 \\
% & Tempo      & 0.37  & 1.58 \\
% & Speaker-only VC  & 65.61 & -12.49 \\
% & Speaker-prosody VC & \textbf{104.43}& \textbf{7.20}\\
% \bottomrule
% \end{tabular}
% % }
% \end{table}

\begin{table}[t]
\caption{Ratio of generated data classified as dysarthria (\%). All, Word, Sent. refers to training data materials for XGBoost.}
% Note that None refers to FLEURS data without augmentation.}
\vspace{-2mm}
\label{tab:ratio}
\centering
\resizebox{0.9\linewidth}{!}{%
\begin{tabular}{cllrr}
\toprule
\textbf{Language} & \textbf{Aug. Method} & \textbf{All} & \textbf{Word} & \textbf{Sent.} \\
\midrule
\multirow{5}{*}{Spanish} 
& None (FLEURS)   &  35.80 & 13.04 & 50.51 \\
\cline{2-5}
& Speed     &42.00 & 15.95 & 52.94 \\
& Tempo   & 36.82  & 14.08 & 50.73 \\
& Speaker-only VC& \textbf{64.35} &20.50  & \textbf{92.90}\\
& Speaker-prosody VC & 59.17 & \textbf{30.62} & 91.10\\
\midrule
\multirow{5}{*}{Italian} 
& None (FLEURS)    & 74.87 & 76.13 & 89.27\\
\cline{2-5}
& Speed   &  73.54  & 75.71 & 88.66\\
& Tempo  &  75.36  & 76.88 & 88.80\\
& Speaker-only VC & 77.39& 78.14 & \textbf{94.14}\\
& Speaker-prosody VC& \textbf{81.64}& \textbf{86.00} & 90.29\\
\midrule
\multirow{5}{*}{Tamil} 
& None (FLEURS)   &  60.51 & 8.04  & 62.25\\
\cline{2-5}
& Speed      & 61.29 & 7.71 & 63.57 \\
& Tempo    & 61.23 &8.07  & 63.24 \\
& Speaker-only VC & 57.00 &  15.89 & 54.93\\
& Speaker-prosody VC & \textbf{69.75} &\textbf{25.61}& \textbf{66.90}\\
\bottomrule
\end{tabular}
}
\vspace{-6mm}
\end{table}

\begin{table}[]
\caption{Perceptual evaluation across different methods.}
\vspace{-2mm}
% ``Dys.?'' indicates a question about how similar the audio sounds to dysarthria, while ``Your Lang.?'' refers to how similar the audio sounds to the listener's native language.}
\label{tab:perceptual}
\centering
\resizebox{0.9\linewidth}{!}{%
\begin{tabular}{clrr}
\toprule
\bf Lang. & \bf Dataset/Aug. Method & \bf Dys.? & \bf Your Lang.? \\
\midrule
\multirow{6}{*}{Spanish} 
& Healthy Data      & 2.50 & --- \\
& Dysarthria Data  & 3.00 & --- \\
\cmidrule{2-4}
& Speed.      & 2.10 & 3.45 \\
& Tempo    & 1.90 & \textbf{3.50} \\
& Speaker-only VC & 2.80 & 2.50 \\
& Speaker-prosody VC & \textbf{3.10} & 1.70 \\
\midrule
\multirow{6}{*}{Italian} 
& Healthy Data     & 1.90 & --- \\
& Dysarthria Data  & 2.50 & --- \\
\cmidrule{2-4}
& Speed       & 1.15 & \textbf{4.00} \\
& Tempo       & 1.10 & \textbf{4.00}\\
& Speaker-only VC  & 1.90 & 3.30 \\
& Speaker-prosody VC  & \textbf{2.45} & 1.70 \\
\midrule
\multirow{6}{*}{Tamil}
& Healthy Data     & 1.00 & --- \\
& Dysarthria Data   & 2.73  & --- \\
\cmidrule{2-4}
& Speed       & 1.60 & \textbf{3.95} \\
& Tempo       & 1.45 & 3.85 \\
& Speaker-only VC  & 2.75 & 2.40 \\
& Speaker-prosody VC & \textbf{3.40} & 2.20 \\
\bottomrule
\end{tabular}
}
\vspace{-6mm}
\end{table}

% \begin{figure}[ht]
%     \centering
%     \begin{minipage}{\linewidth}
%         \centering
%         \includegraphics[width=\textwidth]{figures/eta_divergence_comparison.pdf}
%         \\ 
%     \end{minipage}
% \end{figure}

\subsection{Subjective Evaluation: Perceptual Tests}\label{ssec:perceptual Evaluations}

For the subjective evaluations, we selected 10 audio samples from each group: healthy and dysarthric speech from the dysarthric datasets, and generated speech from each augmentation method. The samples were balanced across severity levels and randomly selected within each.
For each language, two native speakers with no prior training in speech pathology provided judgments in two tasks.
We provided information on perceptual characteristics of dysarthric speech before evaluation\textsuperscript{1}.
The first evaluation was conducted on dysarthric datasets, where participants rated whether the audio sounded dysarthric (``Dys.?'') using a 1-to-4 Likert scale \cite{likert1932technique}, ranging from ``definitely healthy'' to ``definitely dysarthric.''
As the second evaluation, we asked the participants to apply the same scale to generated speech samples, asking how closely generated speech approximates the dysarthric or healthy speech characteristics. For the generated samples, we asked additional question on the naturalness in the target language (``Your Lang.?''). 1-to-4 Likert scale was used, ranging from ``Not at all my language'' to ``Definitely my language.'' This helps to provide information on how well the linguistic characteristics were preserved by the augmentation methods.

\Cref{tab:perceptual} shows that perceptual tests yield similar trends to objective evaluation. Speaker-prosody VC consistently obtained the highest dysarthria ratings, indicating a strongest shift toward dysarthric traits. Speaker-only VC alone also increased dysarthria perception but to a lesser degree, while speed and tempo-based data augmentations had minimal impact relative to healthy data.
Despite its effectiveness in introducing dysarthric characteristics, speaker-prosody VC showed the lowest linguistic similarity ratings, indicating reduced naturalness. By contrast, speed and tempo-based augmentations best maintained linguistic characteristics, with speaker-only VC providing an intermediate balance. These findings underscore a trade-off between enhancing dysarthric characteristics and preserving linguistic naturalness, highlighting the need to balance augmentation strategies according to specific application requirements.
For VC methods, multiple evaluators mentioned, ``They sound like Americans trying to speak my language,'' implying future directions to mitigate such artifacts.

\section{Conclusion}
This study explored VC for dysarthric speech generation in non-English languages, addressing the unavailability of dysarthric data in those languages. 
% imbalance between English and non-English dysarthric speech datasets. 
Experiments in Spanish, Italian, and Tamil demonstrated that augmentation using generated dysarthric speech by VC enhances ASR performances. Objective and subjective evaluations confirmed that our approach effectively simulates dysarthria, highlighting its potential for generating dysarthric speech in low-resource languages and developing inclusive ASR.

% as a scalable augmentation method for low-resource dysarthric ASR.

Despite its promise, our approach has several limitations. This study employed a single VC model as a representative technique; evaluating a broader range of models could uncover key factors to better generate dysarthric-like speech. Additionally, while we examined three languages, expanding the application to a broader array of languages is also necessary. 
% Future work may investigate linguistic factors, such as phonetic similarity and prosodic alignment, to understand their impact on transferring dysarthric characteristics across languages.
% Lastly, 
% Future work may explore multilingual zero-shot prosody transfer to better model the cross-lingual interaction between dysarthric and prosodic features during voice conversion. 
%
% \todo{used UASpeech words as source; usage of sentences.}

% A key challenge is prosodic differences across languages, which can affect the transferability of dysarthric characteristics from English to other languages. 

% By addressing these challenges, cross-lingual VC can serve as a scalable solution for augmenting dysarthric speech data, helping bridge the gap in ASR performance between English and underrepresented languages, and ultimately improving accessibility for individuals with speech impairments worldwide.

\bibliographystyle{IEEEtran}
\bibliography{mybib}

\end{document}